\documentclass[10pt,twocolumn,letterpaper]{article}

\usepackage{cvpr}
\usepackage{times}
\usepackage{epsfig}
\usepackage{graphicx}
\usepackage{amsmath}
\usepackage{amssymb}
\usepackage{subfigure}
\usepackage{array,multirow}
\usepackage{subfigure}
\usepackage{bbding}
\usepackage{booktabs}
\usepackage{threeparttable}
\usepackage{color}
\usepackage{footnote}
\usepackage{microtype}
\def\ie{{\em i.e.}}
\def\eg{{\em e.g.}}
\def\etal{{\em et al.}}

\usepackage[breaklinks=true,bookmarks=false]{hyperref}

\cvprfinalcopy 

\def\cvprPaperID{1782} 

\ifcvprfinal\fi
\begin{document}
	
	\title{ScratchDet: Training Single-Shot Object Detectors from Scratch}
	\author{Rui Zhu$^{1,4}$\thanks{Equally-contributed and this work was done at JD AI Research.},\ \ Shifeng Zhang$^{2*}$, Xiaobo Wang$^1$, Longyin Wen$^3$, Hailin Shi$^1$\thanks{Corresponding author.},\ \ Liefeng Bo$^{3}$, Tao Mei$^{1}$ \\
		$^1$JD AI Research, China; $^2$CASIA, UCAS, China; $^3$JD Digits, USA; $^4$Sun Yat-sen University, China\\
		{\tt\small \textls[-30]{\{zhurui10,wangxiaobo8,longyin.wen,shihailin,liefeng.bo,tmei\}@jd.com, shifeng.zhang@nlpr.ia.ac.cn}}}
	\maketitle
	\thispagestyle{empty}
	
	\begin{abstract}
		Current state-of-the-art object objectors are fine-tuned from the off-the-shelf networks pretrained on large-scale classification dataset ImageNet, which incurs some additional problems: 1) The classification and detection have different degrees of sensitivity to translation, resulting in the learning objective bias; 2) The architecture is limited by the classification network, leading to the inconvenience of modification. To cope with these problems, training detectors from scratch is a feasible solution. However, the detectors trained from scratch generally perform worse than the pretrained ones, even suffer from the convergence issue in training. In this paper, we explore to train object detectors from scratch robustly. By analysing the previous work on optimization landscape, we find that one of the overlooked points in current trained-from-scratch detector is the BatchNorm. Resorting to the stable and predictable gradient brought by BatchNorm, detectors can be trained from scratch stably while keeping the favourable performance independent to the network architecture. Taking this advantage, we are able to explore various types of networks for object detection, without suffering from the poor convergence. By extensive experiments and analyses on downsampling factor, we propose the Root-ResNet backbone network, which makes full use of the information from original images. Our ScratchDet achieves the state-of-the-art accuracy on PASCAL VOC 2007, 2012 and MS COCO among all the train-from-scratch detectors and even performs better than several one-stage pretrained methods. Codes will be made publicly available at \url{https://github.com/KimSoybean/ScratchDet}.
	\end{abstract}
	
	\section{Introduction}
	
	Object detection has made great progress in the framework of convolutional neural networks (CNNs). The current state-of-the-art detectors are generally fine-tuned from high accuracy classification networks, \eg, VGGNet \cite{vgg}, ResNet \cite{res} and GoogLeNet \cite{googlev1} pretrained on ImageNet \cite{imagenet} dataset. The fine-tuning transfers the classification knowledge learned from the source domain to handle the object detection task. In general, fine-tuning from pretrained networks can achieve better performance than training from scratch.
	
	However, there is no such thing as a free lunch. Fine-tuning pretrained networks to object detection has some critical limitations. On the one hand, the classification and detection tasks have different degrees of sensitivity to translation. The classification task prefers to translation invariance, and thus needs downsampling operations (\eg, max-pooling and convolution with stride $2$) for better performance. In contrast, the local texture information is more critical for object detection, making the usage of translation-invariant operations (\eg, downsampling operations) with caution. On the other hand, it is inconvenient to change the architecture of networks (even small changes) in fine-tuning process. If we employ a new architecture, the pretraining should be re-conducted on the large-scale dataset (\eg, ImageNet), requiring high computational cost.
	
	Fortunately, training detectors from scratch is able to eliminate the aforementioned limitations. DSOD \cite{dsod} is the first to train CNN detectors from scratch, in which the deep supervision plays a critical role. Deep supervision is introduced in DenseNet \cite{densely} as the dense layer-wise connection. However, DSOD is also limited by the predefined architecture of DenseNet. If DSOD employs other types of network (\eg, VGGNet and ResNet), the performance decreases dramatically (sometimes even crashes in training). Besides, the currently best performance of trained-from-scratch detectors still remains in a lower place compared with the pretrained ones. Therefore, if we hope to take advantage of training detectors from scratch, it needs to achieve two improvement: (1) free the architecture limitations for any type of network while guarantee the training convergence, (2) give performance as good as pretrained networks (or even better).
	
	To this end, we study the elements that make major impact to the optimization of detector given the randomly initialized network. As pointed out in \cite{santurkar2018does}, BatchNorm reparameterizes the optimization problem to make its landscape significantly smoother instead of reducing the internal covariate shift. Based on this theory, we assume that the lack of BatchNorm in training detector from scratch is the main reason for poor convergence. Thus, we integrate BatchNorm into both the backbone and detection head subnetworks (Figure \ref{fig:structure}), and find that BatchNorm helps the detector converge well in any form of network (including VGGNet and ResNet) without pretraining and surpass the accuracy of the pretrained baselines. Thereby, we are free to modify the architecture without restrictions from pretrained models. By taking this advantage, we analyze the performance of the ResNet and VGGNet based SSD\cite{ssd} detectors  with various configurations, and discover that the sampling stride in the first convolution layer has a great impact on detection performance. Based on this point, we redesign the architecture of the detector by introducing a new root block, which keeps the abundant information for detection feature maps and substantially improves the detection accuracy, especially for small objects. We report extensive experiments on PASCAL VOC 2007 \cite{pascal-voc-2007}, PASCAL VOC 2012 \cite{pascal-voc-2012} and MS COCO \cite{lin2014microsoft} datasets, to demonstrate that our ScratchDet performs better than some pretrained based detectors and all the state-of-the-art train-from-scratch detectors, \eg, improving the state-of-the-art mAP by $1.7\%$ on VOC 2007, $1.5\%$ on VOC 2012, and $2.7\%$ of AP on COCO.
	
	The main contributions of this paper are summarized as follows. (1) We present a single-shot object detector trained from scratch, named ScratchDet, which integrates BatchNorm to help the detector converge well from scratch, independent to the type of network. (2) We introduce a new Root-ResNet backbone network based on the new designed root block, which noticeably improves the detection accuracy, especially for small objects. (3) ScratchDet performs favourably against the state-of-the-art train-from-scratch detectors and some pretrained based detectors.
	
	\begin{figure*}
		\centering
		\subfigure[Loss Value]{
			\label{fig:gradientaaa}
			\includegraphics[width=0.325\linewidth,height=0.325\linewidth]{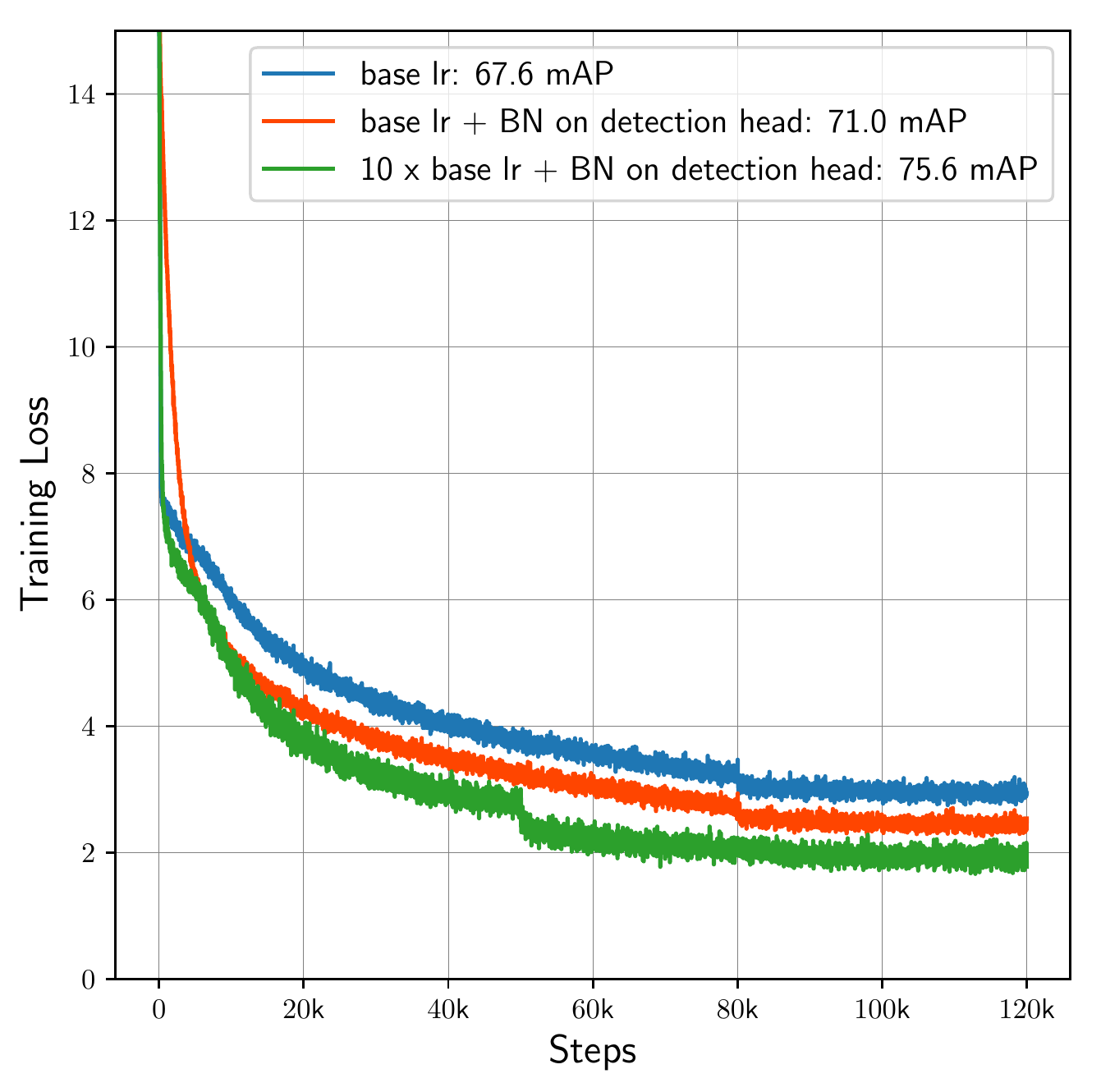}}
		\subfigure[L2 Norm of Gradient]{
			\label{fig:gradientbbb}
			\includegraphics[width=0.325\linewidth,height=0.325\linewidth]{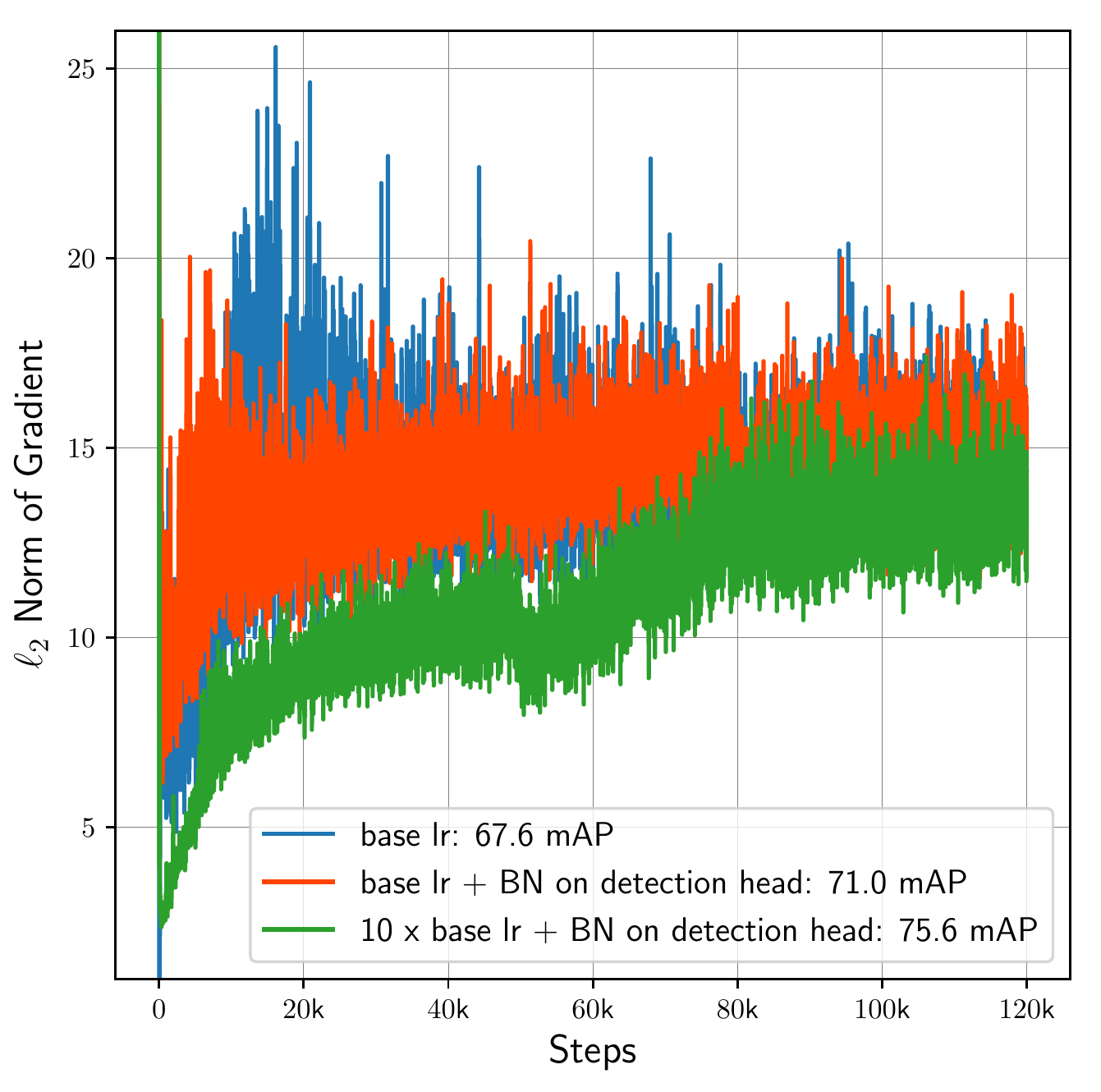}}
		\subfigure[Fluctuation of L2 Norm of Gradient]{
			\label{fig:gradientccc}
			\includegraphics[width=0.325\linewidth,height=0.325\linewidth]{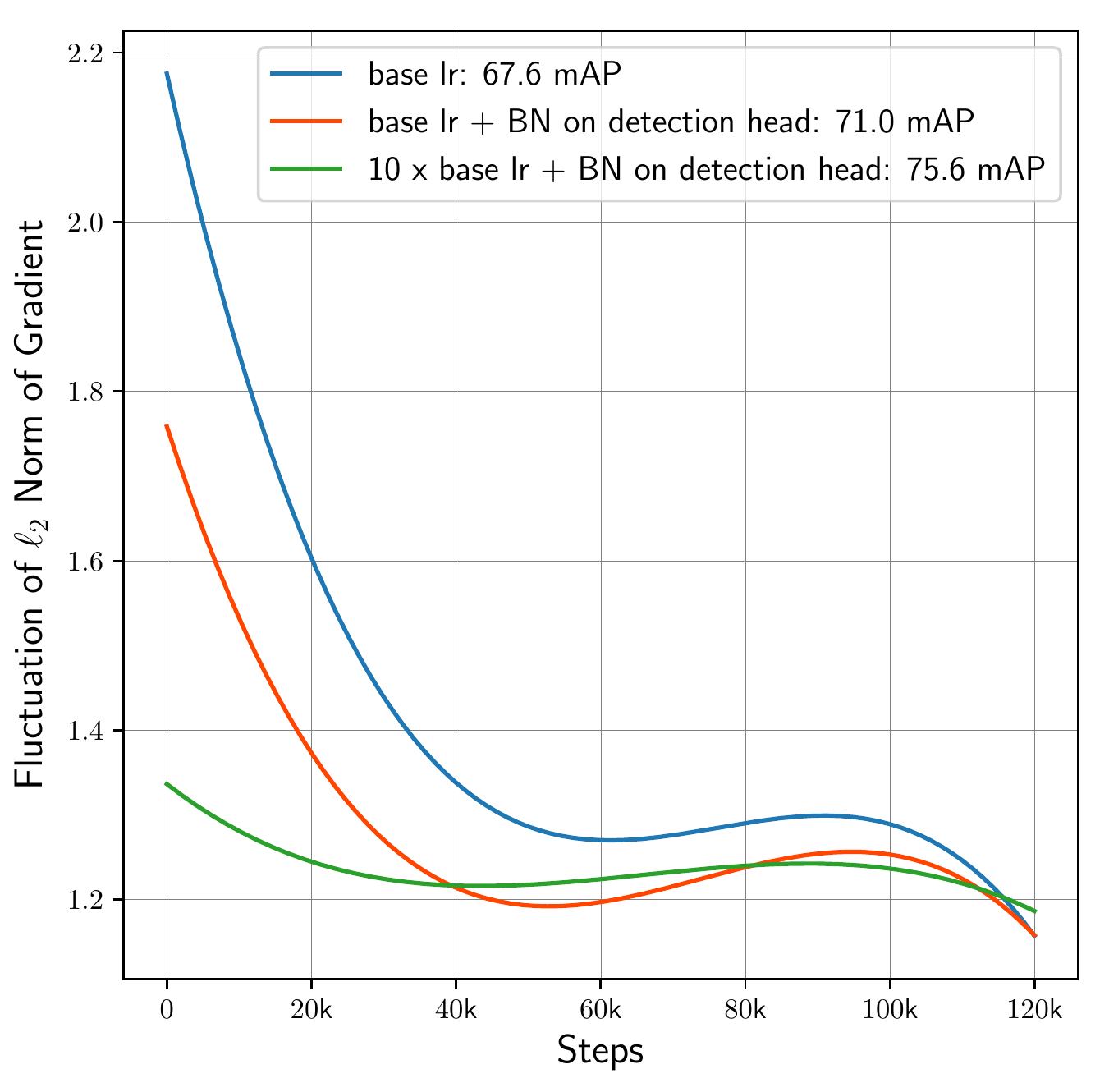}}
		\caption{Optimization landscape analysis. (a) The training loss value. (b) L2 Norm of gradient. (c) Fluctuation of L2 Norm of gradient (smoothed). Blue curve is the original SSD, red and green curves represent the SSD trained with BatchNorm in head subnetwork with $1\times$ and $10\times$ base learning rate, respectively. The BatchNorm makes smoother optimization landscape and has more stable gradients (red \emph{v.s} blue). With this advantage, we are able to set larger learning rate (green) to search larger space and converge faster, and thus better solution.}
		\label{fig:gradient}
	\end{figure*}
	
	\section{Related Work}
	
	{\noindent \textbf{Object detectors with pretrained network.}} Most of CNN-based object detectors are fine-tuned from pretrained networks on ImageNet. Generally, they can be divided into two categories: the two-stage and the one-stage approach. The two-stage approach first generates a set of candidate object proposals, and then predicts the accurate object regions and the corresponding class labels. With the gradual improvements from Faster R-CNN \cite{faster}, R-FCN \cite{RFCN}, FPN \cite{fpn} to Mask R-CNN \cite{he2017mask}, the two-stage methods achieve top performance on several challenging datasets, \eg, PASCAL VOC and MS COCO. Recent developments of two-stage approach focus on redesigning architecture diagram \cite{detnet}, convolution form \cite{dai2017deformable}, re-ranking detection scores \cite{cheng2018revisiting}, using contextual reasoning \cite{bell2016inside} and exploiting multiple layers for prediction \cite{kong2016hypernet}.
	
	Pursuing high efficiency, the one-stage approach attracts much attention in recent years, which simultaneously regresses the object locations and sizes, and the corresponding class labels. OverFeat \cite{sermanet2013overfeat} is one of the first one-stage detectors and since then, several other methods have been
	proposed, such as YOLO \cite{YOLO,yolo9000} and SSD \cite{ssd}. Recent researches on one-stage approach focus on enriching features for detection \cite{fu2017dssd}, designing different architecture \cite{zhang2018single} and addressing class imbalance issue \cite{zhang2017s,focal,zhang2019single}.
	
	{\noindent \textbf{Train-from-scratch object detectors.}} DSOD \cite{dsod} first trains the one-stage object detector from scratch and presents a series of principles to produce good performance. GRP-DSOD \cite{shen2017learning} improves the DSOD algorithm by applying the Gated Recurrent Feature Pyramid. These two methods focus on deep supervision of DenseNet but lose sight of the effect of BatchNorm on optimization and the flexibility of network architecture for training detectors from scratch.
	
	{\noindent \textbf{Batch normalization.}} BatchNorm\cite{batchnorm} addresses the internal covariate shift problem by normalizing layer inputs, which makes using large learning rate to accelerate network training feasible. More recently, Santurkar \etal~\cite{santurkar2018does} provides both empirical demonstration and theoretical justification for the explanation that BatchNorm makes the optimization landscape significantly smoother instead of reducing internal covariate shift. 
	
	
	\section{ScratchDet}
	In this section, we first study the effectiveness of BatchNorm for training SSD from scratch. Then, we redesign the backbone network by analyzing the detection performance of the ResNet and VGGNet based SSD.
	
	\subsection{BatchNorm for train-from-scratch}
	Without losing generality, we consider to apply BatchNorm in SSD which is the most common framework of one stage. SSD is formed by the backbone subnetwork (\eg, truncated VGGNet-16 with several additional convolution blocks) and the detection head subnetwork (\ie, the prediction blocks after each detection layer, which consists of one $3\times3$ bounding box regression convolution layer and one $3\times3$ class label prediction convolution layer). Notice that there is no BatchNorm in the original SSD framework. Motivated by recent work \cite{santurkar2018does}, we believe that using BatchNorm is helpful to train SSD from scratch. BatchNorm makes the optimization landscape significantly smoother, inducing a more predictable and stable behaviour of the gradients to allow for larger searching space and faster convergence. DSOD successfully trains detectors from scratch, however, it attributes the results to deep supervision of DenseNet without emphasizing the effect of BatchNorm. We believe that it is necessary to study the impact of BatchNorm on training detectors from scratch. To verify our argument, we train SSD from scratch using batch size $128$ without BatchNorm as our baseline. As listed in the first column of Table \ref{tab:batchnorm_analysis}, our baseline produces $67.6\%$ mAP on VOC 2007 {\tt test} set.
	
	{\noindent \textbf{BatchNorm in the backbone subnetwork.}} We add BatchNorm in each convolution layer in the backbone subnetwork and then train it from scratch. As shown in Table \ref{tab:batchnorm_analysis}, using BatchNorm in the backbone network improves $5.2\%$ of mAP. More importantly, adding BatchNorm in the backbone network makes the optimization landscape significantly smoother. Thus, we can use larger learning rates ($0.01$ and $0.05$) to further improve the performance (\ie, mAP is improved from $72.8\%$ to $77.8\%$ and $78.0\%$). Both of them outperform SSD fine-tuned from the pretrained VGG-16 model ($77.2\%$ \cite{ssd}). These results indicate that adding BatchNorm in the backbone subnetwork is one of the critical issues to train SSD from scratch.
	
	{\noindent \textbf{BatchNorm in the detection head subnetwork.}} To analyze the effect of BatchNorm in the detection head subnetwork, we plot the training loss value, L2 Norm of gradient, and fluctuation of L2 Norm of gradient \emph{v.s} training steps. As shown by the blue curve in Figure \ref{fig:gradientbbb} and \ref{fig:gradientccc}, training SSD from scratch with default learning rate $0.001$ has a large fluctuation of L2 norm of gradient, especially in the initial phase of training, which makes the loss value suddenly change and converge to a bad local minima (\ie, relatively high loss at the end of training process in Figure \ref{fig:gradientaaa} and bad detection result $67.6\%$ mAP). These results are useful to explain the phenomenon that using large learning rate to train SSD with the original architecture from scratch or pretrained networks usually leads to gradient explosion, poor stability and weak prediction of gradients (see Table \ref{tab:batchnorm_analysis}) . 
	
	In contrast, integrating BatchNorm in the detection head subnetwork makes the loss landscape smoother (see red curves in Figure \ref{fig:gradient}), which improves mAP from $67.6\%$ to $71.0\%$  (listed in Table \ref{tab:batchnorm_analysis}). 
	The smooth landscape allows us to set larger learning rate, which brings about larger searching space and faster convergence (see Figure \ref{fig:gradientaaa} and \ref{fig:gradientccc}). As a result, the mAP improves from $71.0\%$ to $75.6\%$. Besides, with BatchNorm, larger learning rate is also helpful to jump out of the bad local minima and produce stable gradients (green curve in Figure \ref{fig:gradientbbb} and \ref{fig:gradientccc}).

	{\noindent \textbf{BatchNorm in the whole network.}} We also study the performance of the detector using BatchNorm in both the backbone and detection head subnetworks. After using BatchNorm in the whole network of detector, we are able to use a larger base learning rate ($0.05$) to train the detector from scratch, which produces $1.5\%$ higher mAP comparing to the detector initialized with the pretrained VGG-16 backbone ($78.7\%$ \emph{v.s} $77.2\%$). Please see Table \ref{tab:batchnorm_analysis} for more details.
	
	\subsection{Backbone Network}
	As described above, we train SSD with BatchNorm from scratch and achieve better accuracy than the pretrained SSD. This encourages us to train detector from scratch while keeping the performance independent to the network architecture. By taking this advantage, we are able to explore various types of network for the object detection task.
	
\begin{figure}
		\centering
		\includegraphics[width=1.0\linewidth]{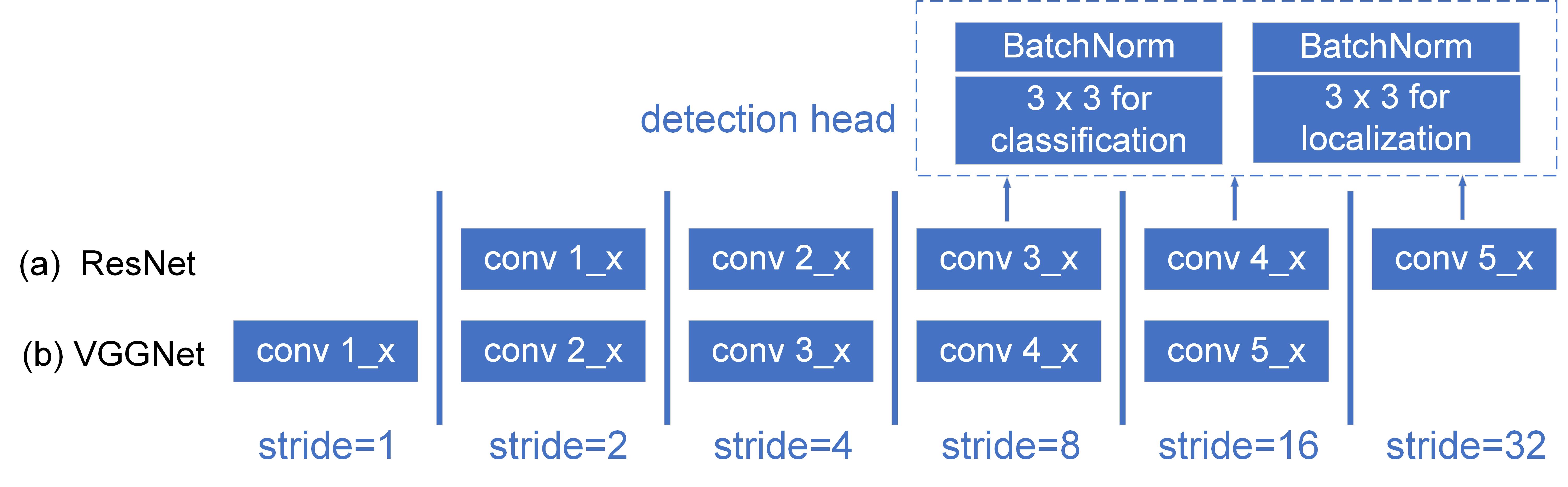}
		\vspace{-5mm}
		\caption{Brief overview of SSD based on VGG-16 and ResNet-101. The BatchNorm is covered for clearness. As shown in Figure \ref{fig:jiegou0} and Table \ref{tab:backbone_analysis}, the first stride 2 of ResNet makes worse performance on PASCAL VOC with small input size.}
		\vspace{-2mm}
		\label{fig:structure}
	\end{figure}

\begin{figure}[!h]
		\centering
		\includegraphics[width=0.43\textwidth, height=0.63\textwidth]{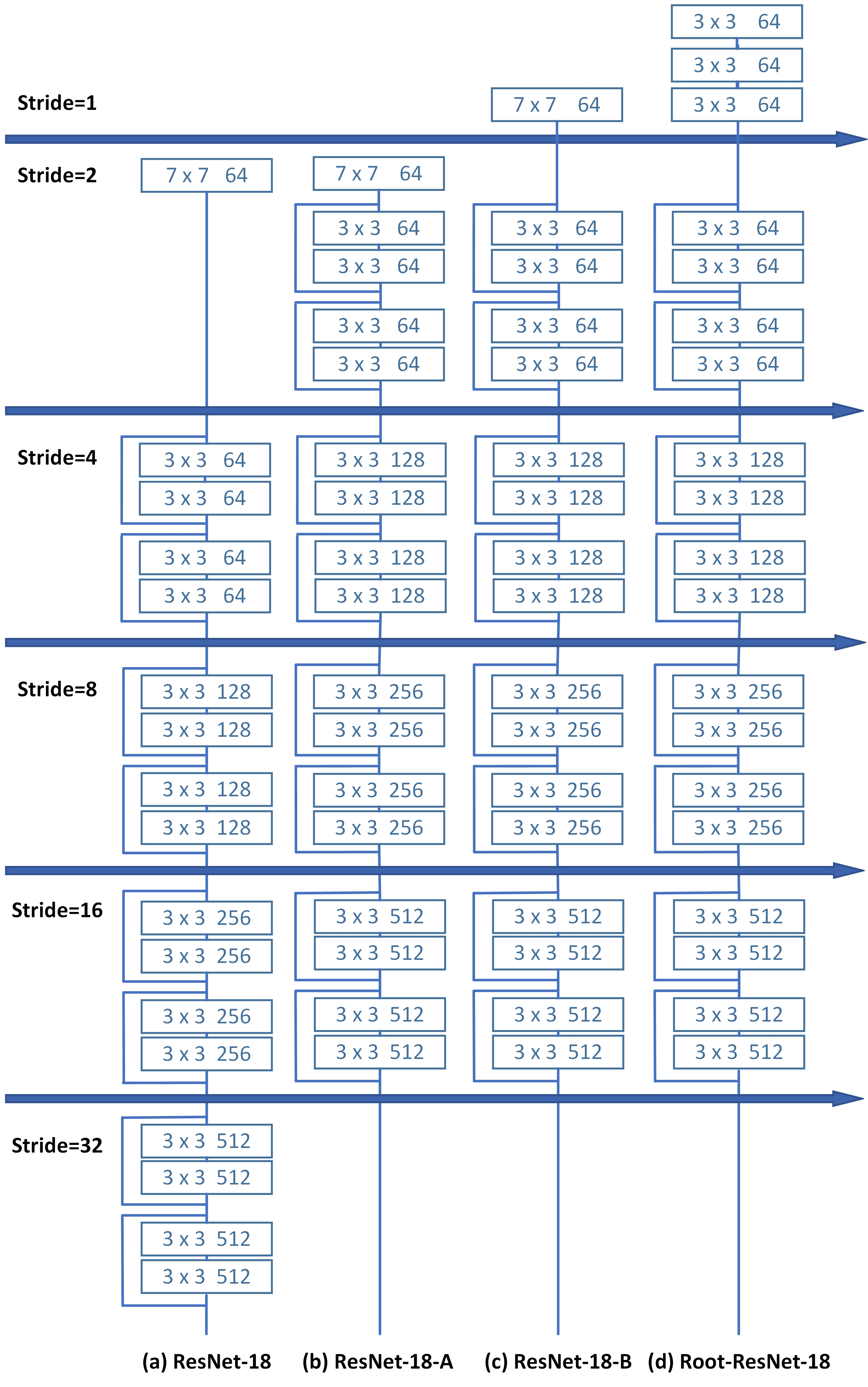}
		\caption{Illustration of networks in Section \ref{Section 4.2.2}. (a) ResNet-18: original structure. (b) ResNet-18-A: removing the first max-pooling layer. (c) ResNet-18-B: changing the stride size in the first conv layer from 2 to 1. (d) Root-ResNet-18: replacing the $7\times7$ conv layer with three stacked $3\times3$ conv layers in ResNet-18-B. The corresponding mAPs on PASCAL 2007 test (training on ``07+12'' from scratch) are $73.1\%$, $75.3\%$, $77.6\%$ and $78.5\%$, respectively. Notably, for a fairy comparison, no matter how we modify the structure, the spatial sizes of our selected detection layers are the same as SSD300 and DSOD300 (\ie, $38\times38$, $19\times19$, $10\times10$, $5\times5$, $3\times3$, $1\times1$).}
		\vspace{-2mm}
		\label{fig:jiegou0}
	\end{figure}

	{\noindent \textbf{Performance analysis of ResNet and VGGNet.}} The truncated VGG-16 and ResNet-101 are two popular backbone networks used in SSD (a brief structure overview in Figure \ref{fig:structure}). In general, ResNet-101 produces better classification results than VGG-16 (\eg, $5.99\%$ \emph{v.s} $8.68\%$, $2.69\%$ top-5 classification error lower on ImageNet). However, as indicated in DSSD \cite{fu2017dssd}, the VGG-16 based SSD performs favourably than the ResNet-101 based SSD with relatively small input size (\eg, $300\times300$) on PASCAL VOC. We argue that this phenomenon is caused by the downsampling operation in the first convolution layer (\ie, conv1\_x with stride $2$) of ResNet-101. This operation significantly affects the detection accuracy, especially for small objects (see Table \ref{tab:backbone_analysis}). After we remove the downsampling operation in conv1\_x of ResNet-18 to form ResNet-18-B in Figure \ref{fig:jiegou0}(c), the detection performance improves by a big margin from $73.1\%$ to $77.6\%$ mAP. We also remove the second downsampling operation to form ResNet-18-A in Figure \ref{fig:jiegou0}(b), whose improvement is relatively small. In summary, the downsampling operation in the first convolution layer has a bad impact on the detection accuracy, especially for small objects. 
	
	{\noindent \textbf{Backbone network redesign for object detection.}} To overcome the disadvantages of ResNet based backbone network for object detection while retaining its powerful classification ability, we design a new architecture, named Root-ResNet, which is an improvement of the truncated ResNet in the original SSD detector, shown in Figure \ref{fig:jiegou0}(d). We remove the downsampling operation in the first conv layer and replace the $7\times7$ convolution kernel by a stack of $3\times3$ convolution filters (similar as the stem block in DSOD\cite{dsod}, but denoted as the root block due to the large influence from the first stride). With abundant inputs, Root-ResNet is able to exploit more local information from the image, so as to extract powerful features for small object detection. Furthermore, we replace the four convolution blocks (added by SSD to extract the feature maps with different scales) with four residual blocks to the end of the Root-ResNet. Each residual block is formed by two branches. One branch is a $1\times1$ convolution layer with stride $2$ and the other one consists of a $3\times3$ convolution layer with stride $2$ and a $3\times3$ convolution layer with stride $1$. The number of output channels in each convolution layer is set to $128$. These residual blocks bring efficiency in parameters and computation without performance dropout. 
	
	\begin{table*}[h]
		\centering
		\caption{Analysis of BatchNorm and learning rate for SSD trained from scratch on VOC 2007 {\tt test} set. All the networks are based on the truncated VGG-16 backbone network. The best performance ($78.7\%$ mAP) is achieved when three conditions are satisfied: (1) BatchNorm in backbone and head, (2) non pretraining, (3) larger learning rate. ``NAN'' indicates that the training is non-convergent.}
		\vspace{1mm}
		\footnotesize \setlength{\tabcolsep}{4.7pt}
		\begin{tabular}{c|ccc ccc| ccc ccc| ccc ccc}
			\hline
			\multicolumn{1}{c|}{Component}&\multicolumn{6}{c|}{lr 0.001}&\multicolumn{6}{c|}{lr 0.01}&\multicolumn{6}{c}{lr 0.05}\\
			\hline
			pretraining & & & & &\Checkmark &\Checkmark & & & & &\Checkmark &\Checkmark &  &  &  &  &\Checkmark  &\Checkmark  \\
			BN in backbone & & &\Checkmark &\Checkmark  &  &\Checkmark  & & &\Checkmark &\Checkmark  &  &\Checkmark  & & &\Checkmark &\Checkmark  &  &\Checkmark  \\
			BN in head & &\Checkmark & &\Checkmark &\Checkmark &\Checkmark &  &\Checkmark  &  &\Checkmark  & \Checkmark &\Checkmark  &  &\Checkmark  &  &\Checkmark  &\Checkmark  &\Checkmark  \\
			\hline
			mAP (\%) & 67.6 & 71.0 & 72.8 & 71.8 & 77.1 & 77.6 & NAN & 75.6 & 77.8 & 77.3 & 76.9 & 78.2 & NAN & NAN & 78.0 &\textcolor{red}{78.7} & NAN & 75.5 \\
			\hline
		\end{tabular}
		\vspace{-3mm}
		\label{tab:batchnorm_analysis}
	\end{table*}
	
	\section{Experiment}
	We conduct several experiments on the PASCAL VOC and MS COCO datasets, including $20$ and $80$ object classes. The proposed ScratchDet is implemented in Caffe library \cite{jia2014caffe} and all the codes and the trained models will be made publicly available.
	
	\subsection{Training details}
	All models are trained from scratch using SGD with $0.0005$ weight decay and $0.9$ momentum on $4$ NVIDIA Tesla P40 GPUs. For a fair comparison, we use the same training settings as the original SSD, including data augmentation, anchor settings and loss function. We remove the L2 normalization \cite{liu2015parsenet}. Notably, all experiments select the detection layers with the fixed spatial size same as SSD300 and DSOD300, \ie, do not use larger-size feature maps for detection. Following DSOD, we use a relatively large batch size $128$ to train our ScratchDet from scratch, in order to ensure the stable statistical results of BatchNorm in training phase. Meanwhile, we use the default batch size $32$ for the pretrained model based SSD (We also try $128$ batch size for the pretrained model, but the performance has not improved).
	
	Notably, we use the ``Root-ResNet-18'' redesigned from ResNet-18 as the backbone network in the model analysis by considering the computational cost in experiments. Whereas, in comparison with the state-of-the-art detectors, we use a deeper backbone network ``Root-ResNet-34'' for better performance. All the parameters in our ScratchDet are initialized by the ``xavier'' method \cite{glorot2010understanding}. Besides, all the models are trained with the $300\times300$ input size and we believe that the accuracy of ScratchDet can be further improved using larger input size.

	\subsection{PASCAL VOC 2007}
	For PASCAL VOC 2007, all models are trained on the VOC 2007 and VOC 2012 {\tt trainval} sets ($16,551$ images), and tested on the VOC 2007 {\tt test} set ($4,952$ images). We use the same settings and configurations except for some specified changes of model components.
	
	\subsubsection{Analysis of BatchNorm} We construct several variants of the original SSD and evaluate them on VOC 2007 to demonstrate the effectiveness of BatchNorm in training SSD from scratch, shown in Table \ref{tab:batchnorm_analysis}.
	
	{\noindent\textbf{Without BatchNorm.}}
	We train the original SSD from scratch with the batch size $128$. All the other settings are the same as that in \cite{ssd}. As shown in the first column of Table \ref{tab:batchnorm_analysis}, we get $67.6\%$ mAP, which is $9.6\%$ worse than the detector initialized by the pretrained classification network (\ie, $77.2\%$). In addition, due to the unstable gradient and unsmooth optimization landscape, the training is able to successfully converge only with the learning rate $0.001$ and goes to a bad local minima (see blue curves in Figure \ref{fig:gradient}). As shown in Table \ref{tab:batchnorm_analysis}, if we use larger learning rates ($0.01$ and $0.05$), the training process will not converge.
	
	{\noindent \textbf{BatchNorm in the backbone subnetwork.}}
	BatchNorm is a widely used to enable fast and stable training of deep neural networks. To validate the effectiveness of BatchNorm in the backbone subnetwork, we add the BatchNorm operation to each convolution layer in the truncated VGG-16 network, denoted as VGG-16-BN, and train the VGG-16-BN model based SSD from scratch. As shown in Table \ref{tab:batchnorm_analysis}, using BatchNorm in the backbone network with relative large learning rate ($0.05$) improves mAP from $67.6\%$ to $78.0\%$.
	
	{\noindent \textbf{BatchNorm in the detection head subnetwork.}}
	We also study the effectiveness of BatchNorm in the detection head subnetwork. As described before, the detection head subnetwork in SSD is used to predict the locations, sizes and class labels of objects. The original SSD method \cite{ssd} do not use BatchNorm in detection head subnetwork. As presented in Table \ref{tab:batchnorm_analysis}, we find that using BatchNorm only on the detection head subnetwork improves $3.4\%$ mAP from $67.6\%$ to $71.0\%$. After using the $10$ times larger base learning rate $0.01$, the performance can be further improved from $71.0\%$ to $75.6\%$. This noticeable improvement ($8.0\%$) demonstrates the importance of using BatchNorm in the detection head subnetwork.
	
	{\noindent 	\textbf{BatchNorm in the whole network.}}
	We use BatchNorm on every convolution layer in SSD and train it from scratch with three different base learning rates ($0.001$, $0.01$ and $0.05$). For the $0.001$ and $0.01$ base learning rates, we achieve $71.8\%$ and $77.3\%$ mAPs, respectively. When we use the largest learning rate $0.05$, the performance will be further improved by $1.4\%$ mAP to $78.7\%$, which outperforms the pretrained network based SSD detector ($78.7\%$ \emph{v.s} $77.2\%$). These results indicate that using BatchNorm on each convolution layers in SSD is critical to train it from scratch.
	
	{\noindent 	\textbf{BatchNorm for the pretrained network.}}
	To validate the effect of BatchNorm for SSD finetuning from pretrained networks, we construct a variant of the original SSD, \ie, adding the BatchNorm operation to every convolution layer. The layers in backbone network are initialized by the pretrained VGG-16-BN model from ImageNet, which is converted from the PyTorch official model. As shown in Table \ref{tab:batchnorm_analysis}, we observe that the best result achieves $78.2\%$ with learning rate $0.01$. Comparing to the original SSD fine-tuned from the pretrained network, BatchNorm improves only $1.0\%$ mAP ($77.2\%$ \emph{v.s} $78.2\%$) of the detector, which is rather small compared to the improvement of the trained-from-scratch detector (\ie, $11.1\%$ mAP improvement from $67.6\%$ to $78.7\%$)\footnote{we also try the batch size $128$ with default settings of SSD, producing $78.2\%$ mAP for VGG-16-BN and $76.8\%$ mAP for VGG-16 without improvement.}. We would also like to emphasize that ScratchDet produces better performance than the BatchNorm based SSD trained from the pretrained network (\ie, $78.7\%$ \emph{v.s} $78.2\%$). The results demonstrate that BatchNorm is more critical for SSD trained from scratch than fine-tuned from pretrained models.
	
	{\noindent \textbf{BatchNorm in DSOD. }}
	DSOD attributes its success to deep supervision of DenseNet and ignores the effect of BatchNorm. After removing all BatchNorm layers in DSOD, the mAP drops $6.2\%$ from $77.7\%$ to $71.5\%$ on VOC 2007. Thus, we argue BatchNorm rather than deep supervision is the key to train detectors form scratch and experiments in Table \ref{tab:batchnorm_analysis}  validate this point. Besides, training VGG16-based Faster R-CNN without BatchNorm from scratch cannot converge in the DSOD paper, but with BatchNorm it can converge successfully to $67.2\%$ mAP, although it is still lower than the pretrained one ($73.2\%$ mAP).
	
	\subsubsection{Analysis of the backbone subnetwork.}
	\label{Section 4.2.2}
	We analyze the pros and cons of the ResNet and VGGNet based SSD detectors and redesign the backbone network, called Root-ResNet. Specifically, all the models are designed based on the ResNet-18 backbone network in experiments. We also use BatchNorm in the detection head subnetwork. In the training phase, the learning rate is set to $0.05$ for the first $45k$ iterations, and is divided by $10$ successively for another $30k$, $20k$ and $5k$ iterations, respectively. As shown in Table \ref{tab:backbone_analysis}, training SSD from scratch based on ResNet-18 only produces $73.1\%$ mAP. We analyze the reasons as follows.
	
	{\noindent \textbf{Kernel size in the first layer.}}
	In contrast to VGG16, the first convolution layer in ResNet-18 uses relatively large kernel size $7\times7$ with stride $2$. We aim to explore the effect of the kernel size of the first convolution layer on the detector trained from scratch. As shown in the first two rows of Table \ref{tab:backbone_analysis}, the kernel size of convolution layer has no impact on the performance (\ie, $73.1\%$ for $7\times7$ {\em v.s} $73.2\%$ for $3\times3$). Using smaller kernel size $3\times3$ produces a slightly better results with faster speed. The same conclusion can be obtained when we set the stride size of the first convolution layer to $1$ without downsampling, see the fifth and the sixth row of Table \ref{tab:backbone_analysis} for more details.
	
	{\noindent  \textbf{Downsampling in the first layer.}}
	Compared to VGGNet, ResNet-18 uses downsampling on the first convolution layer, leading to considerable local information loss, which greatly impacts the detection performance, especially for small objects. As shown in Table \ref{tab:backbone_analysis}, after removing the downsampling operation in the first layer (\ie, ResNet-18-B in Figure \ref{fig:jiegou0}), we can improve $4.5\%$ and $4.6\%$ mAPs for the $7\times7$ and $3\times3$ kernel sizes, respectively. When we only remove the second downsampling operation and keep the first stride = 2 (\ie, ResNet-18-A in Figure \ref{fig:jiegou0}), the performance achieves $75.3\%$ mAP,  $2.3\%$ lower than modifying the first layer ($77.6\%$ mAP). These results demonstrate that the downsampling operation in the first convolution layer is the obstacle for good results. We need to remove this operation when training ResNet based SSD from scratch.
	
	{\noindent  \textbf{Number of layers in the root block.}}
	Inspired by DSOD and GoogLeNet-V3 \cite{szegedy2016rethinking}, we use several convolution layers with kernel size $3\times3$ to replace the $7\times7$ convolution layers (\ie, Root-ResNet-18 in Figure \ref{fig:jiegou0}). Here, we study the impact of number of stacked convolution layers in the root block on the detection performance in Table \ref{tab:backbone_analysis}. As the number of convolution layers increasing from $1$ to $3$, the mAP scores are improved from $77.8\%$ to $78.5\%$. However, the accuracy decreases as the number of stacked layers becoming larger than $3$. We believe that three $3\times3$ convolution layers in the root block are enough to learn the information from raw images, and adding more $3\times3$ layers cannot boost the accuracy any more. Empirically, we use three $3\times3$ convolution layers for detection task on PASCAL VOC 2007, 2012 and MS COCO datasets with $300\times300$ input size.
	
	The aforementioned conclusions can be also extended to deeper ResNet backbone network, \eg, ResNet-34. As shown in Table \ref{tab:pascal-voc}, using Root-ResNet-34, the mAP of our ScratchDet is improved from $78.5\%$ to $80.4\%$, which is the best results with $300\times300$ input size. In comparison experiments on the benchmarks, we use Root-ResNet-34 as the backbone network.

	\begin{table}[t]
		\centering
		\caption{Analysis of backbone network for SSD trained from scratch on VOC 2007 {\tt test} set. All models are based on the ResNet-18 backbone. FPS is measured on one Tesla P40 GPU.}
		\vspace{1mm}
		\footnotesize \setlength{\tabcolsep}{12.0pt}
		\begin{tabular}{c|c|c|c}
			\hline
			First conv layer & Root block & FPS & mAP\\
			\hline
			\multirow{3}*{with downsmapling} & {\bf 1}: 7$\times$7 &59.5 &73.1\\	
			& {\bf 1}: 3$\times$3 &62.9 &73.2\\
			& {\bf 2}: 3$\times$3 &58.1 &74.9\\
			& {\bf 3}: 3$\times$3 &54.5 &75.4\\
			\hline
			\multirow{6}*{without downsmapling} & {\bf 1}: 7$\times$7 & 37.0 & 77.6\\
			& {\bf 1}: 3$\times$3 & 37.2 & 77.8\\
			& {\bf 2}: 3$\times$3 & 31.5 & 78.1\\
			& {\bf 3}: 3$\times$3 & 26.9 & \bf{78.5}\\
			& {\bf 4}: 3$\times$3 & 24.3 & 78.4\\
			& {\bf 5}: 3$\times$3 & 21.8 & 78.5\\
			\hline
		\end{tabular}
		\label{tab:backbone_analysis}
	\end{table}
	
	\begin{table*}[t]
		\centering
		\caption{Detection results on the PASCAL VOC datasets. For VOC 2007, all methods are trained on the VOC 2007 and 2012 {\tt trainval} sets and tested on the VOC 2007 {\tt test} set. For VOC 2012, all methods are trained on the VOC 2007 and 2012 {\tt trainval} sets plus the VOC 2007 {\tt test} set, and tested on the VOC 2012 {\tt test} set. The FPS of ScratchDet is measured on one TITAN X GPU for the fair comparison. $^\dagger$: {\scriptsize \url{http://host.robots.ox.ac.uk:8080/anonymous/0HPCHC.html}} $^\ddagger$: {\scriptsize \url{http://host.robots.ox.ac.uk:8080/anonymous/JSL6ZY.html}}  } 
		\vspace{1mm}
		\footnotesize \setlength{\tabcolsep}{4.5pt}
		\begin{tabular}{p{4.0cm}<{\centering}|p{3.0cm}<{\centering}|p{2.5cm}<{\centering}|p{1.5cm}<{\centering}|p{2.2cm}<{\centering}|p{2.2cm}<{\centering}}
			\hline
			\multirow{2}{*}{Method} &\multirow{2}{*}{Backbone} &\multirow{2}{*}{Input size} &\multirow{2}{*}{FPS} &\multicolumn{2}{c}{mAP (\%)} \\
			\cline{5-6}
			& & & &VOC 2007 &VOC 2012 \\
			\hline
			\hline
			\textit{pretrained two-stage:} & & & & & \\
			HyperNet \cite{kong2016hypernet} &VGG-16 &$\sim1000\times600$ &0.88 &76.3 &71.4\\
			Faster R-CNN\cite{faster} &ResNet-101 &$\sim1000\times600$ &2.4 &76.4 & 73.8\\
			ION\cite{bell2016inside}  &VGG-16 &$\sim1000\times600$ &1.25 &76.5 &76.4\\
			MR-CNN\cite{gidaris2015object}  &VGG-16 &$\sim1000\times600$ &0.03 &78.2 &73.9\\
			R-FCN\cite{RFCN} &ResNet-101 &$\sim1000\times600$ &9 &80.5 &77.6\\
			CoupleNet\cite{zhu2017couplenet} &ResNet-101 &$\sim1000\times600$ &8.2 &82.7 &80.4\\
			\hline
			\textit{pretrained one-stage:} & & & & & \\
			RON384\cite{kong2017ron} &VGG-16 &$384\times384$ &15 &74.2 &71.7\\
			SSD321\cite{fu2017dssd} &ResNet-101 &$321\times321$ &11.2 &77.1 &75.4\\
			SSD300$^*$\cite{ssd}& VGG16 &$300\times300$ &46 &77.2 &75.8\\
			YOLOv2\cite{yolo9000}  &Darknet-19 &$544\times544$ &40 &78.6 &73.4\\
			DSSD321\cite{fu2017dssd} &ResNet-101 &$321\times321$ &9.5 &78.6 &76.3\\
			DES300\cite{zhang2018des}    &VGG-16     &$300\times300$ &29.9 &79.7 &77.1\\
			RefineDet320\cite{zhang2018single}    &VGG-16     &$320\times320$ &40.3 &80.0 &78.1\\
			\hline
			\hline
			\textit{trained from scratch:} & & & & & \\
			DSOD300\cite{dsod} &DS/64-192-48-1 &$300\times300$ &17.4 &77.7 &76.3\\
			GRP-DSOD320\cite{shen2017learning} &DS/64-192-48-1 &$300\times300$ &16.7 &78.7 &77.0\\
			ScratchDet300    &Root-ResNet-34     &$300\times300$ &17.8 &80.4 &78.5$^\dagger$\\
			ScratchDet300+   &Root-ResNet-34     &- &- &\bf{84.1} &\bf{83.6}$^\ddagger$ \\
			\hline
		\end{tabular}
		\vspace{-3mm}
		\label{tab:pascal-voc}
	\end{table*}
	
	\subsubsection{Results}
	
	We compare ScratchDet to the state-of-the-art detectors in Table \ref{tab:pascal-voc}. With small input $300\times300$, ScratchDet produces $80.4\%$ mAP without bells and whistles, better than several state-of-the-art one-stage pretrained object detectors (\eg, $80.0\%$ mAP of RefineDet320 and $79.7\%$ mAP of DES300). Note that we keep most of original SSD configurations and the same epochs with DSOD. The result is much better than SSD300-VGG16 ($80.4\%$ \emph{v.s} $77.2\%$ and $3.2\%$ mAP higher) and SSD321-ResNet101 ($80.4\%$ \emph{v.s} $77.1\%$, $3.3\%$ mAP higher). ScratchDet outperforms the state-of-the-art train-from-scratch detector with $1.7\%$ improvements on mAP score (\ie, $80.4\%$ \emph{v.s} $78.7\%$ of GRP-DSOD). In the multi-scale testing, our ScratchDet achieves $84.1\%$ (ScratchDet300+) mAP, which is the state-of-the-art.
	
	\subsection{PASCAL VOC 2012}
	Following the evaluation protocol of VOC 2012, we use VOC 2012 {\tt trainval} set, and VOC 2007 {\tt trainval} and {\tt test} sets ($21,503$ images) to train our ScratchDet from scratch, and test on VOC 2012 {\tt test} set ($10,991$ images). The detection results of ScratchDet are submitted to the public testing server for evaluation. The learning rate and batch size are set the same as that in VOC 2007.
	
	Table \ref{tab:pascal-voc} reports the accuracy of ScratchDet as well as the state-of-the-art methods. Using small input size $300\times300$, ScratchDet produces $78.5\%$ mAP, surpassing some one-stage methods with similar input size, \eg, SSD321-ResNet101 ($75.4\%$, $3.1\%$ higher mAP), DES300-VGG16 ($77.1\%$, $1.4\%$ higher mAP), and RefineDet320-VGG16 ($78.1\%$, $0.4\%$ higher mAP). Meanwhile, comparing to the two-stage methods based on pretrained networks with $\sim1000\times600$ input size, ScratchDet also produces better results than R-FCN ($77.6\%$, $0.9\%$ higher mAP). In addition, our ScratchDet outperforms all the train-from-scratch detectors. It outperforms DSOD by $2.2\%$ mAP with $60$ less training epochs and surpasses GRP-DSOD by $1.5\%$ mAP. Notably, in the multi-scale testing, ScratchDet obtains $83.6\%$ mAP, much better than the state-of-the-arts of both one-stage and two-stage methods.
	
	\begin{table*}[t]
		\centering
		\caption{Detection results on the MS COCO {\tt test-dev} set.}
		\vspace{1mm}
		\footnotesize \setlength{\tabcolsep}{10pt}
		\begin{tabular}{c|c|c|ccc|ccc}
			\hline
			Method &Data &Backbone &AP &AP$_{50}$ &AP$_{75}$ &AP$_{\it S}$ &AP$_{\it M}$ &AP$_{\it L}$\\
			\hline
			\hline
			\textit{pretrained two-stage:} & & & & & & & & \\
			ION\cite{bell2016inside} &train &VGG-16 &23.6 &43.2 &23.6 &6.4 &24.1 &38.3\\
			OHEM++ \cite{shrivastava2016training}&trainval &VGG-16 &25.5 &45.9 &26.1 &7.4 &27.7 &40.3 \\
			R-FCN\cite{RFCN} &trainval &ResNet-101 &29.9 &51.9 &- &10.8 &32.8 &45.0\\
			CoupleNet\cite{zhu2017couplenet} &trainval &ResNet-101 &34.4 &54.8 &37.2 &13.4 &38.1 &50.8 \\
			Faster R-CNN+++ \cite{res}&trainval &ResNet-101-C4 &34.9 &55.7 &37.4 &15.6 &38.7 &50.9\\
			Faster R-CNN w FPN \cite{fpn}&trainval35k &ResNet-101-FPN &36.2 &59.1 &39.0 &18.2 &39.0 &48.2 \\
			Faster R-CNN w TDM\cite{shrivastava2016beyond} &trainval &Inception-ResNet-v2-TDM &36.8 &57.7 &39.2 &16.2 &39.8 &52.1 \\
			Deformable R-FCN\cite{dai2017deformable} &trainval &Aligned-Inception-ResNet &37.5 &58.0 &40.8 &19.4 &40.1 &52.5 \\
			Mask R-CNN\cite{he2017mask} &trainval35k &ResNet-101-FPN &38.2 &{\bf60.3} &41.7 &20.1 &43.2 &{\bf51.2} \\
			\hline
			\textit{pretrained one-stage:} & & & & & & & & \\
			YOLOv2\cite{yolo9000} &trainval35k &DarkNet-19 &21.6 &44.0 &19.2 &5.0 &22.4 &35.5\\
			
			SSD300$^*$\cite{ssd} &trainval35k &VGG16 &25.1 &43.1 &25.8 &6.6 &25.9 &41.4\\
			RON384++\cite{kong2017ron} &trainval &VGG-16 &27.4 &49.5 &27.1 &- &- &- \\
			SSD321\cite{fu2017dssd} &trainval35k &ResNet-101 &28.0 &45.4 &29.3 &6.2 &28.3 &49.3\\
			DSSD321\cite{fu2017dssd} &trainval35k &ResNet-101 &28.0 &46.1 &29.2 &7.4 &28.1 &47.6\\
			DES300\cite{zhang2018des} &trainval35k &VGG-16 &28.3 &47.3 &29.4 &8.5 &29.9 &45.2 \\
			DFPR300 \cite{kong2018deep}  &trainval &VGG-16 &28.4 &48.2 &29.1 &8.2 &30.1 &44.2\\
			RefineDet320\cite{zhang2018single}  &trainval35k &VGG-16 &29.4 &49.2 &31.3 &10.0 &32.0 &44.4\\
			DFPR300 \cite{kong2018deep} &trainval &ResNet-101 &31.3 &50.5 &32.0 &10.5 &33.8 &49.9 \\
			PFPNet-R320 \cite{kim2018parallel} &trainval35k &VGG-16 &31.8 &52.9 &33.6 &12.0 &35.5 &46.1 \\
			RetinaNet400\cite{focal} &trainval35k &ResNet-101 &31.9 &49.5 &34.1 &11.6 &35.8 &48.5 \\
			RefineDet320\cite{zhang2018single}   &trainval35k &ResNet-101 &32.0 &51.4 &34.2 &10.5 &34.7 &50.4 \\
			\hline
			\hline
			\textit{trained from scratch:} & & & & & & & & \\
			DSOD300\cite{dsod} &trainval  &DS/64-192-48-1 &29.3 &47.3 &30.6 &9.4 &31.5 &47.0\\
			GRP-DSOD320\cite{shen2017learning} &trainval  &DS/64-192-48-1 &30.0 &47.9 &31.8 &10.9 &33.6 &46.3\\
			ScratchDet300 &trainval35k &Root-ResNet-34 &32.7 &52.0 &34.9 &13.0 &35.6 &49.0 \\
			ScratchDet300+ &trainval35k &Root-ResNet-34 &{\bf 39.1} &59.2 &{\bf 42.6} &{\bf 23.1} &{\bf 43.5} &51.0 \\
			\hline
		\end{tabular}
		\label{tab:coco}
	\end{table*}
	
	\subsection{MS COCO}
	We also evaluate ScratchDet on MS COCO dataset. The model is trained from scratch on the MS COCO {\tt trainval35k} set and tested on the {\tt test-dev} set. We set the base learning rate to $0.05$ for the first $150k$ iterations, and divide it by $10$ successively for another $100k$, $60k$ and $10k$ iterations respectively.
	
	Table \ref{tab:coco} shows the results on the MS COCO {\tt test-dev} set. ScratchDet produces $32.7\%$ AP that is better than all the other methods with similar input size by a large margin, such as SSD300 ($25.1\%$, $7.6\%$ higher AP), SSD321 ($28.0\%$, $4.7\%$ higher AP), GRP-DSOD320 ($30.0\%$, $2.7\%$ higher AP), DSSD321 ($28.0\%$, $4.7\%$ higher AP), DES300 ($28.3\%$, $4.4\%$ higher AP), RefineDet320-VGG16 ($29.4\%$, $3.3\%$ higher AP), RetinaNet400 ($31.9\%$, $0.8\%$ higher AP) and RefineDet320-ResNet101 ($32.0\%$, $0.7\%$ higher AP). Notably, with the same input size, DSOD300 trains on the {\tt trainval} set, which contains $5000$ more images than {\tt trainval35k} (\ie, $123,287$ {\em v.s} $118,287$), and our ScratchDet produces a much better result ($32.7\%$ {\em v.s} $29.3\%$, $3.4\%$ higher AP). Some methods use much bigger input sizes for both training and testing ($\sim1000\times600$) than our ScratchDet300, \eg, CoupleNet, Faster R-CNN and Deformable R-FCN. For a fair comparison, we also report the multi-scale testing AP results of ScratchDet300 in Table \ref{tab:coco}, \ie, $39.1\%$, which is currently the best result, surpassing those prominent two-stage and one-stage approaches with large input image sizes.  
	
	Comparing to the state-of-the-art methods with similar input image size, ScratchDet300 produces the best AP$_{\it S}$ ($13.0\%$) for small objects, outperforming SSD321 by $6.8\%$. The significant improvement in small object demonstrates the superiority of our ScratchDet architecture for small object detection.

	\begin{table}[t]
		\centering
		\caption{Detection results on PASCAL VOC dataset. All models are pretrained on MS COCO, and fine-tuned on PASCAL VOC. $^\dagger$: {\scriptsize \url{http://host.robots.ox.ac.uk:8080/anonymous/ZVCMYN.html}} $^\ddagger$: {\scriptsize \url{http://host.robots.ox.ac.uk:8080/anonymous/OFHUPV.html}}}  
		\vspace{1mm}
		\footnotesize \setlength{\tabcolsep}{4pt}
		\begin{tabular}{c|c|c|c}
			\hline
			\multirow{2}{*}{Method}   &\multirow{2}{*}{Backbone} &\multicolumn{2}{c}{mAP (\%)} \\
			\cline{3-4}
			& & VOC 2007 & VOC 2012\\
			\hline
			\hline
			\textit{pretrained two-stage:} & & &\\
			Faster R-CNN\cite{faster}  &VGG-16 &78.8 &75.9   \\
			OHEM++\cite{shrivastava2016training}    &VGG-16 &- &80.1 \\
			R-FCN\cite{RFCN}     &ResNet-101 &83.6 &82.0\\
			\hline
			\textit{pretrained one-stage:} & & &\\
			SSD300\cite{ssd}    &VGG-16         &81.2 &79.3 \\
			RON384++\cite{kong2017ron}  &VGG-16         &81.3 &80.7 \\
			RefineDet320\cite{zhang2018single}   &VGG-16        &84.0      &82.7 \\
			\hline
			\hline
			\textit{trained without ImageNet:} & & & \\
			DSOD300\cite{dsod}   &DS/64-192-48-1 &81.7 &79.3 \\
			ScratchDet300   &Root-ResNet-34      &84.0      &82.1$^\dagger$ \\
			ScratchDet300+ &Root-ResNet-34 &{\bf 86.3} &{\bf 86.3$^\ddagger$}  \\
			\hline
		\end{tabular}
		\label{tab:coco-to-voc}
	\end{table}

	\subsection{From MS COCO to PASCAL VOC}
	We also study how the MS COCO dataset help the detection on PASCAL VOC. Since the object classes in PASCAL VOC are from an subset of MS COCO, we directly fine-tune the detection models pretrained on MS COCO by subsampling parameters. As shown in Table \ref{tab:coco-to-voc}, ScratchDet300 achieves $84.0\%$ and $82.1\%$ mAP on the VOC 2007 {\tt test} set and VOC 2012 {\tt test} set, outperforming other train-from-scratch methods. In the multi-scale testing, the detection accuracies are promoted to $86.3\%$ and $86.3\%$, respectively. By using the training data in MS COCO and PASCAL VOC, our ScratchDet obtains the top mAP scores on both VOC 2007 and 2012 datasets.
	
	\subsection{Comparison of the training time}
	ScratchDet uses obviously more time than fine-tuning a pretrained classifier on the 4 NVIDIA Tesla P40 GPUs workstation with the $300 \times 300$ input image size for the MS COCO dataset (\ie, $84.6$ vs. $29.7$ hours). However, considering several weeks and millions of images involved in the pretraining phase, training detectors from scratch is more attractive than the pretrained detector. Notice that the comparison of training time is based on mmdetection framework \cite{mmdetection2018}.
	
	\section{Conclusion}
	In this work, we focus on training object detectors from scratch in order to tackle the problems caused by fine-tuning from pretrained networks. We study the effects of BatchNorm in the backbone and detection head subnetworks, and successfully train detectors from scratch. By taking the pretaining-free advantage, we are able to explore various architectures for detector designing. After analyzing the performance of the ResNet and VGGNet based SSD, we propose a new Root-ResNet backbone network to further improve the detection accuracy, especially for small objects. As a consequence, the proposed detector sets a new state-of-the-art performance on the PASCAL VOC 2007, 2012 and MS COCO datasets for the train-from-scratch detectors, even outperforming some one-stage pretrained methods.
	
	\section*{Acknowledgements}
	We thank the engineers Jianhao Zhang and Peng Cheng in JD AI Research for their helpful suggestions for this work.
	
	{\small
		\bibliographystyle{ieee_fullname}
		\bibliography{egbib}
	}

\clearpage
\onecolumn
\begin{center}
	\LARGE
	\textbf{Appendix}\\\
	\\
	
\end{center}

\maketitle

\section{Complete Object Detection Results}
\begin{figure*}[!h]
	\centering
	\includegraphics[width=0.98\textwidth]{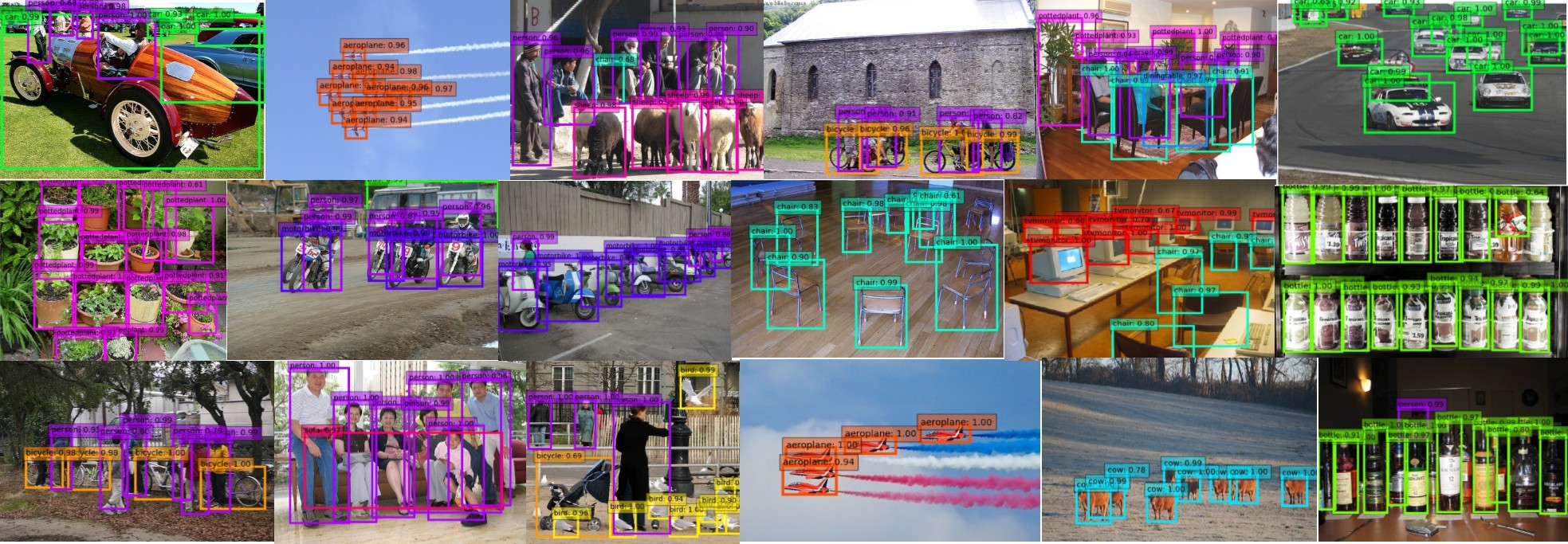}
	\caption{Qualitative results of ScratchDet300 on the PASCAL VOC 2007 {\tt test} set (corresponding to $84.0\%$ mAP). The training data is 07+12+COCO.}
	\label{fig:pascal-voc-2007}
\end{figure*}

\begin{figure*}[!h]
	\centering
	\includegraphics[width=0.98\textwidth]{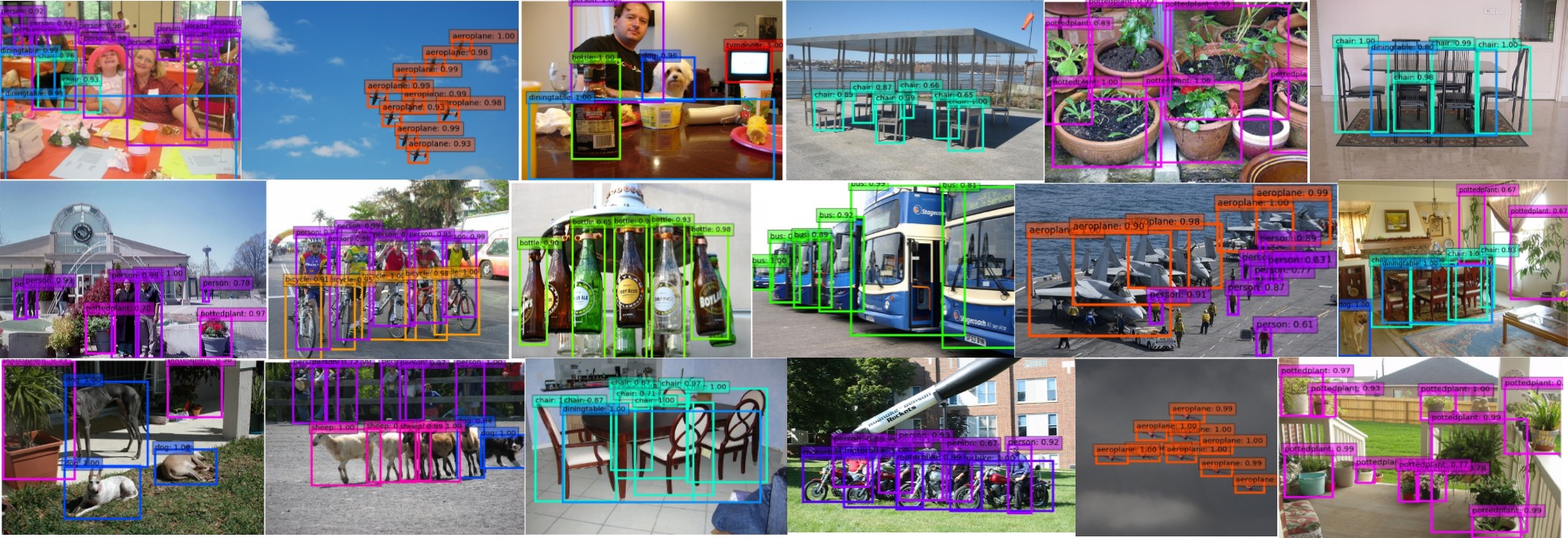}
	\caption{Qualitative results of ScratchDet300 on the PASCAL VOC 2012 {\tt test} set (corresponding to $82.1\%$ mAP). The training data is 07++12+COCO.}
	\label{fig:pascal-voc-2012}
\end{figure*}

\begin{figure*}[!t]
	\centering
	\includegraphics[width=0.98\textwidth]{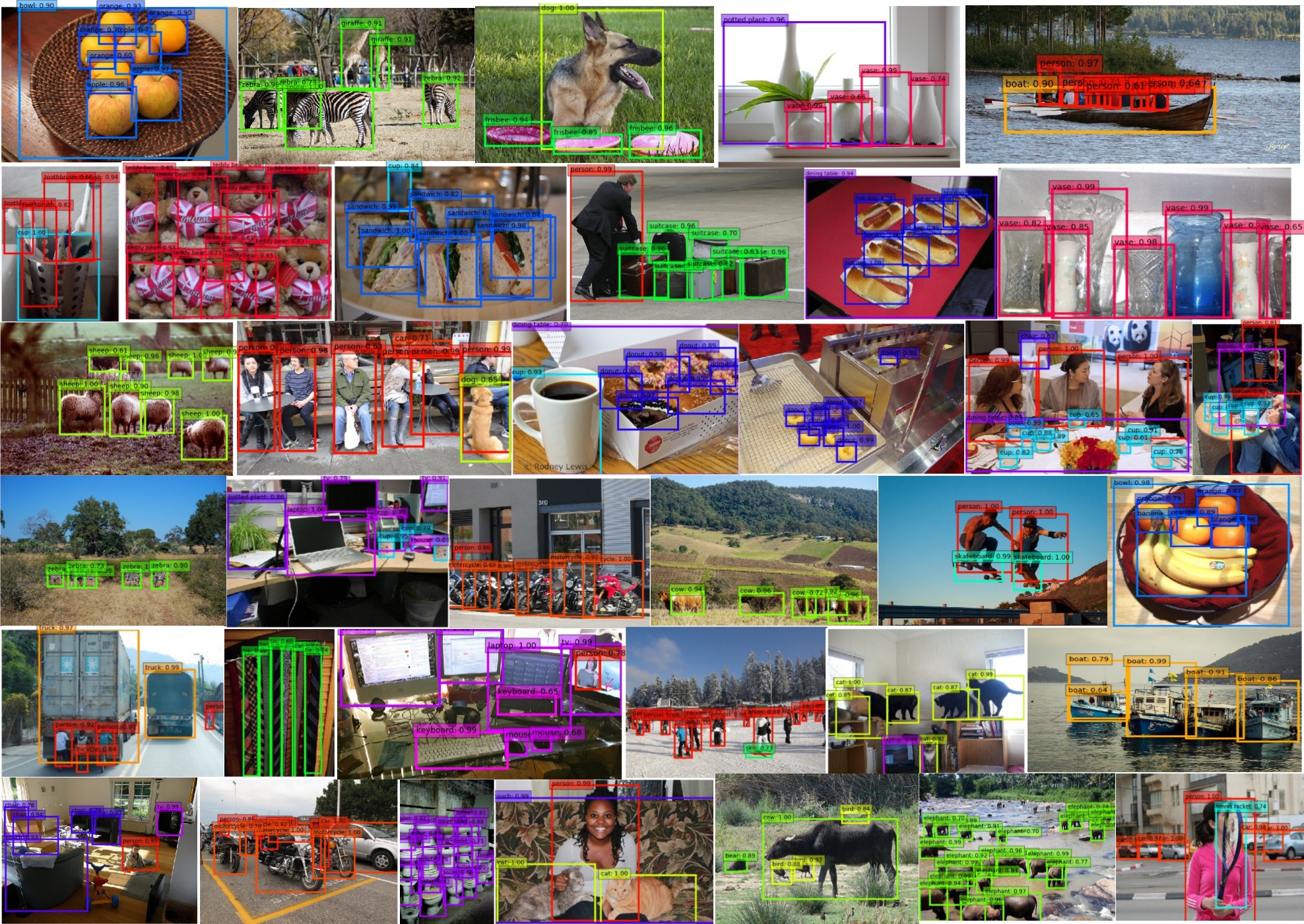}
	\caption{Qualitative results of ScratchDet300 on the MS COCO {\tt test-dev} set (corresponding to $32.7\%$ mAP). The training data is COCO {\tt trainval35k}.}
	\label{fig:coco}
\end{figure*}

We show the complete object detection results of the proposed ScratchDet method on the PASCAL VOC 2007 {\tt test} set, PASCAL VOC 2012 {\tt test} set and MS COCO {\tt test-dev} set in Table \ref{tab:pascal-voc-2007}, Table \ref{tab:pascal-voc-2012} and Table \ref{tab:coco}, respectively. Among the results of all published methods, our ScratchDet achieves the best performance on these three detection datasets, \ie, $86.3\%$ mAP on the PASCAL VOC 2007 {\tt test} set, $86.3\%$ mAP on the PASCAL VOC 2012 {\tt test} set and $39.1\%$ AP on the MS COCO {\tt test-dev} set. And we select some detection examples on the PASCAL VOC 2007 {\tt test} set, the PASCAL VOC 2012 {\tt test} set and the MS COCO {\tt test-dev} in Figure \ref{fig:pascal-voc-2007}, Figure \ref{fig:pascal-voc-2012}, and Figure \ref{fig:coco}, respectively. Different colors of the bounding boxes indicate different object categories. Our method works well with the occlusions, truncations, inter-class interference and clustered background.

\begin{table*}[h]
	\centering
	\caption{Object detection results on the PASCAL VOC 2007 {\tt test} set. All models use Root-ResNet-34 as the backbone network.}
	\footnotesize \setlength{\tabcolsep}{1.5pt}
	\begin{tabular}{c|c|c|cccccccccccccccccccc}
		\toprule[1.5pt]
		Method &Data &mAP &aero &bike &bird &boat &bottle &bus &car &cat &chair &cow &table &dog &horse &mbike &person &plant &sheep &sofa &train &tv \\
		\hline
		ScratchDet300 &07+12 &80.4 &86.0 &87.7 &77.8 &73.9 &58.8 &87.4 &88.4 &88.2 &66.4 &84.3 &78.4 &84.0 &87.5 &88.3 &83.6 &57.3 &80.3 &79.9 &87.9 &81.2\\
		ScratchDet300+ &07+12 &84.1 &90.0 &89.2 &83.6 &80.0 &70.1 &89.3 &89.5 &89.0 &73.0 &86.9 &79.8 &87.4 &90.1 &89.3 &87.1 &63.3 &86.9 &83.5 &88.9 &83.4 \\
		\hline
		ScratchDet300 &COCO+07+12 &84.0 &87.9 &89.3 &85.6 &79.8 &69.4 &89.1 &89.2 &88.5 &73.2 &87.5 &81.7 &88.4 &89.5 &88.7 &86.3 &63.1 &84.5 &84.3 &88.1 &85.6 \\
		ScratchDet300+ &COCO+07+12 &86.3 &90.4 &89.6 &88.4 &85.4 &78.9 &90.1 &89.3 &89.5 &77.4 &89.7 &83.9 &89.1 &90.3 &89.5 &88.3 &68.1 &87.6 &85.9 &87.4 &87.7 \\
		\bottomrule[1.5pt]
	\end{tabular}
	\label{tab:pascal-voc-2007}
\end{table*}

\begin{table*}[h]
	\centering
	\caption{Object detection results on the PASCAL VOC {\tt 2012 test} set. All models use Root-ResNet-34 as the backbone network.}
	\footnotesize \setlength{\tabcolsep}{1.4pt}
	\begin{tabular}{c|c|c|cccccccccccccccccccc}
		\toprule[1.5pt]
		Method &Data &mAP &aero &bike &bird &boat &bottle &bus &car &cat &chair &cow &table &dog &horse &mbike &person &plant &sheep &sofa &train &tv \\
		\hline
		ScratchDet300 &07++12 &78.5 &90.1 &86.8 &74.5 &66.3 &54.0 &83.7 &82.6 &91.6 &64.1 &83.1 &67.7 &90.1 &87.6 &87.8 &85.7 &56.9 &81.7 &74.6 &87.2 &75.3 \\
		ScratchDet300+ &07++12 &83.6 &92.2 &90.3 &82.6 &73.9 &68.1 &86.8 &90.5 &93.9 &70.3 &88.0 &72.3 &92.3 &91.5 &91.0 &90.3 &63.6 &87.6 &77.4 &89.9 &80.2 \\
		\hline
		ScratchDet300 &COCO+07++12 &82.1 &91.7 &89.3 &79.1 &71.9 &62.7 &85.7 &85.3 &93.9 &68.8 &87.2 &68.7 &91.9 &90.6 &90.9 &88.2 &61.2 &84.7 &79.2 &89.7 &81.0 \\
		ScratchDet300+ &COCO+07++12 &86.3 & 94.0 &91.8 &86.0 &78.9 &75.6 &88.6 &91.3 &95.1 &74.0 &90.0 &73.0  &93.6 &93.0 &92.6 &91.9 &69.7 &90.2 &80.9 &91.8 &83.7 \\
		\bottomrule[1.5pt]
	\end{tabular}
	\label{tab:pascal-voc-2012}
\end{table*}

\begin{table*}[!h]
	\centering
	\caption{Object detection results on the MS COCO {\tt test-dev} set. All models use Root-ResNet-34 as the backbone network.}
	\setlength{\tabcolsep}{7.0pt}
	\begin{threeparttable}
		\begin{tabular}{c|ccc|ccc|ccc|ccc}
			\toprule[1.5pt]
			Method  &AP &AP$_{50}$ &AP$_{75}$ &AP$_{\it S}$ &AP$_{\it M}$ &AP$_{\it L}$ &AR$_{1}$ &AR$_{10}$ &AR$_{100}$ &AR$_{\it S}$ &AR$_{\it M}$ &AR$_{\it L}$\\
			\hline
			ScratchDet300  &32.7 &52.2 &34.9 &13.0 &35.6 &49.0 &29.3 &43.9 &45.7 &20.6 &50.8 &65.3\\
			ScratchDet300+ &39.1 &59.2 &42.6 &23.1 &43.5 &51.0 &33.1 &53.3 &58.3 &36.6 &63.4 &74.5 \\
			\bottomrule[1.5pt]
		\end{tabular}
	\end{threeparttable}
	\label{tab:coco}
\end{table*}

\begin{figure*}
	\centering
	\subfigure[Loss Value]{
		\label{fig:gradientaaatbackbonebn}
		\includegraphics[width=0.325\linewidth]{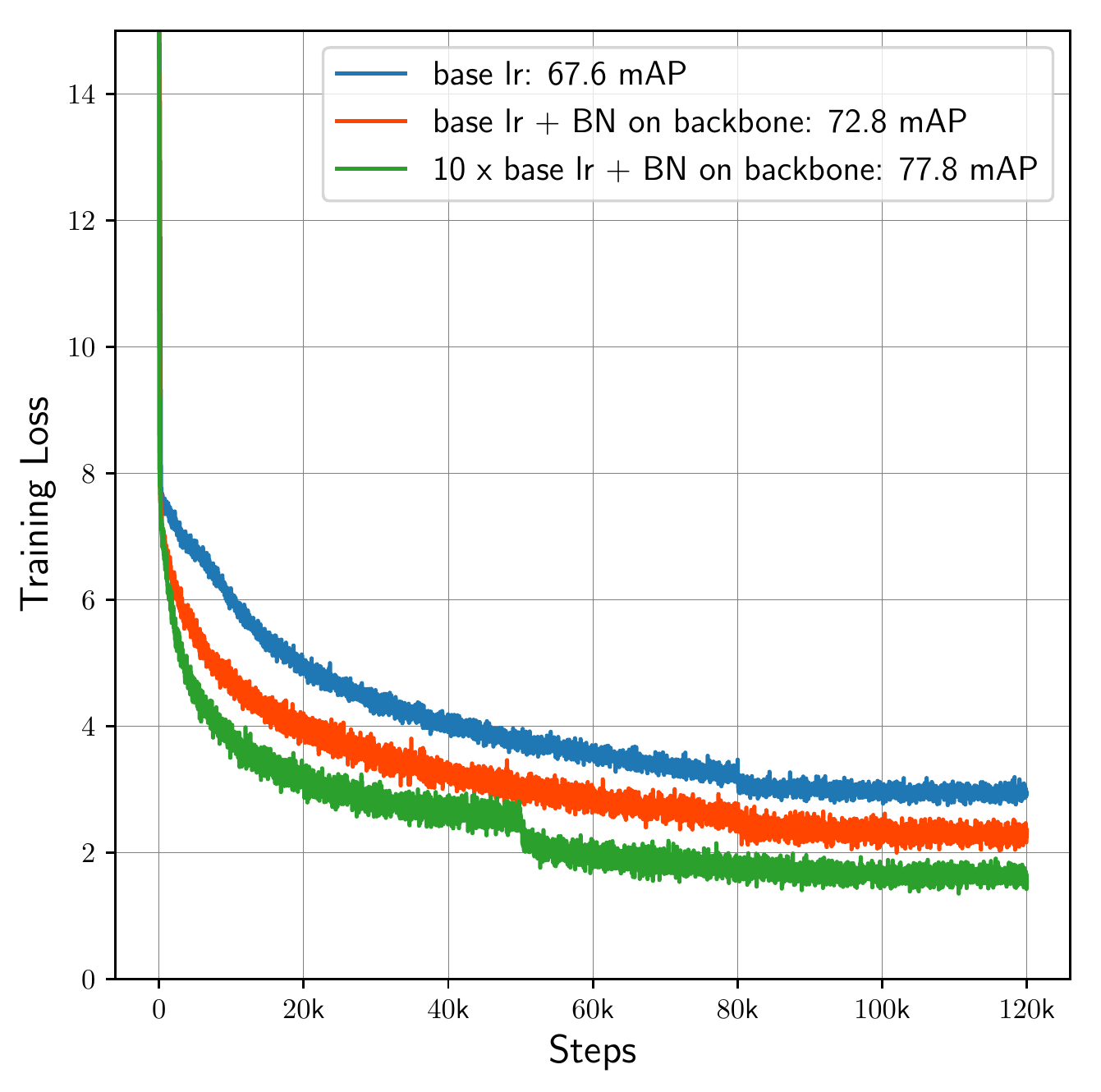}}
	\subfigure[L2 Norm of Gradient]{
		\label{fig:gradientaaatbackbonebn}
		\includegraphics[width=0.325\linewidth]{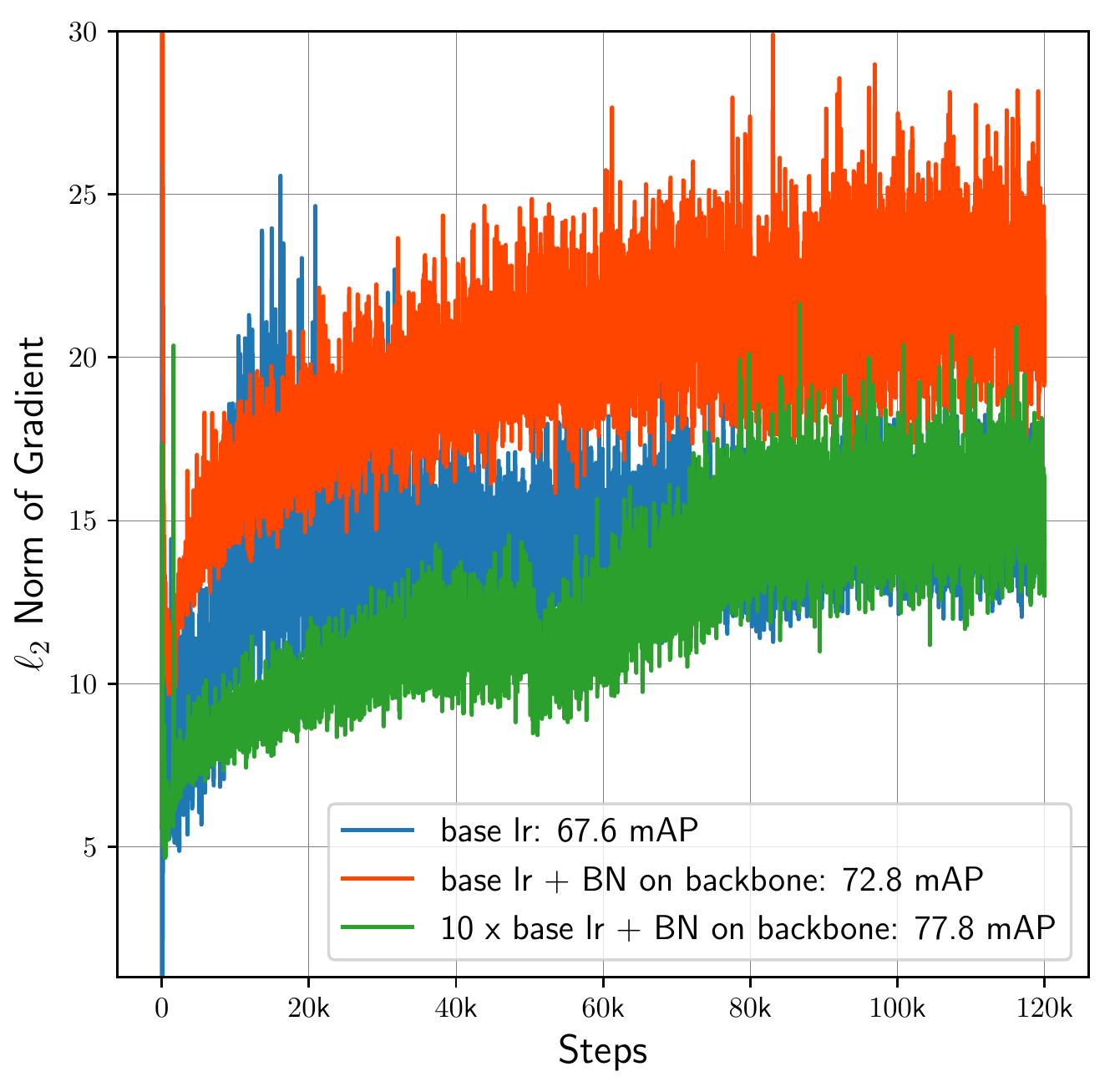}}
	\subfigure[Fluctuation of L2 Norm of Gradient]{
		\label{fig:gradientcccbackbonebn}
		\includegraphics[width=0.325\linewidth]{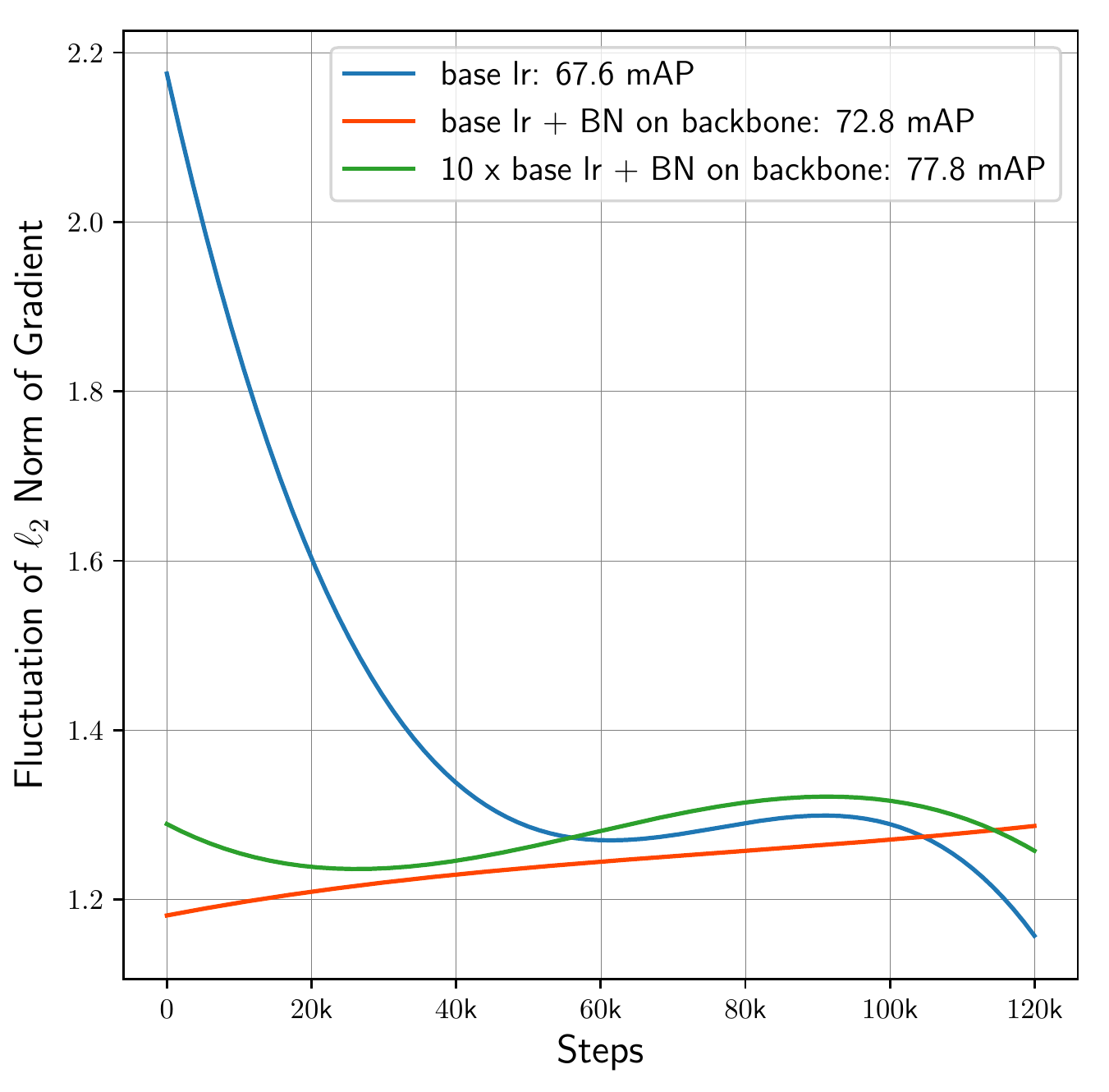}}
	\caption{Analysis of the optimization landscape of SSD after adding BatchNorm on the backbone subnetwork. We plot (a) the training loss value, (b) L2 Norm of gradient and (c) the fluctuation of L2 Norm of gradient of three detectors. The blue curve represents the original SSD, the red and green curves represent the SSD trained with BatchNorm on the backbone network using base learning rate and $10\times$ base learning rate, respectively. It is the similar trend with the curves of adding BatchNorm on the detection head subnetwork.}
	\label{fig:gradientbackbonebn}
\end{figure*}

\begin{figure}
	\centering
	\includegraphics[width=0.5\linewidth, height=0.5\linewidth]{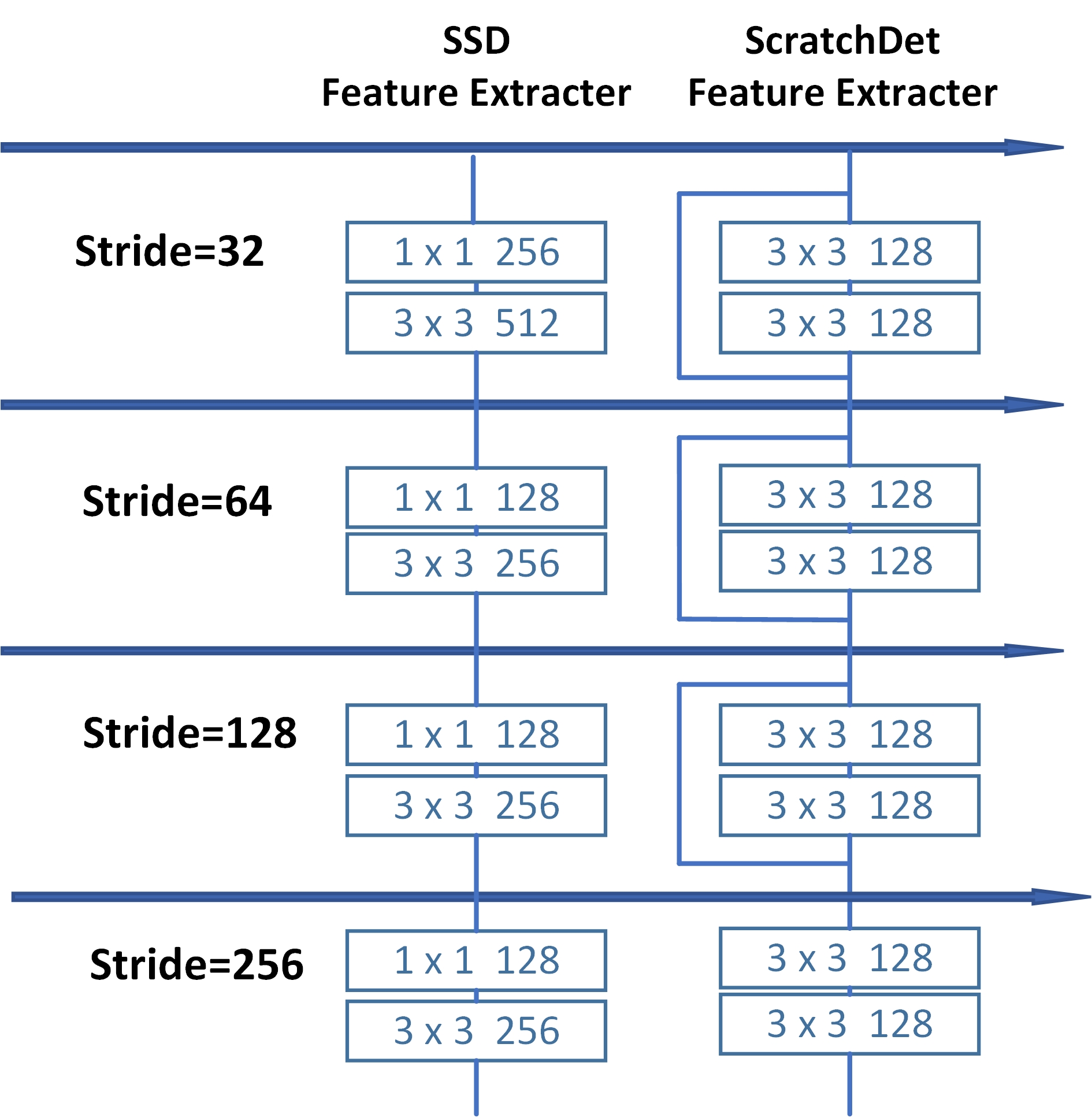}
	\caption{Comparison of the extra added layers between SSD and ScratchDet. This change brings less parameters and computions.}
	\label{fig:jiegou1}
\end{figure}

\end{document}